\documentclass[conference]{IEEEtran}
\IEEEoverridecommandlockouts
\usepackage{cite}
\usepackage{amsmath,amssymb,amsfonts}
\usepackage{algorithmic}
\usepackage{graphicx}
\usepackage{textcomp}
\usepackage{csquotes}
\usepackage{xcolor}
\usepackage{adjustbox}
\usepackage{hyperref}
\usepackage{multirow}
\usepackage{orcidlink}

\def\BibTeX{{\rm B\kern-.05em{\sc i\kern-.025em b}\kern-.08em
    T\kern-.1667em\lower.7ex\hbox{E}\kern-.125emX}}
\DeclareMathOperator{\argmin}{argmin}
\begin{document}

\title{Data-Driven Strategies for Coping with Incomplete DVL Measurements}

\author{\IEEEauthorblockN{Nadav Cohen\IEEEauthorrefmark{1}\orcidlink{0000-0002-8249-0239} and Itzik Klein \orcidlink{0000-0001-7846-0654}}
\IEEEauthorblockA{\textit{The Hatter Department of Marine Technologies} \\
\textit{Charney School of Marine Sciences, University of Haifa}\\
Haifa, Israel}

\thanks{\IEEEauthorrefmark{1}Corresponding author: N. Cohen (email: ncohe140@campus.haifa.ac.il).}}

\maketitle

\begin{abstract}
Autonomous underwater vehicles are specialized platforms engineered for deep underwater operations. Critical to their functionality is autonomous navigation, typically relying on an inertial navigation system and a Doppler velocity log. In real-world scenarios, incomplete Doppler velocity log measurements occur, resulting in positioning errors and mission aborts. To cope with such situations, a model and learning approaches were derived. This paper presents a comparative analysis of two cutting-edge deep learning methodologies, namely LiBeamsNet and MissBeamNet, alongside a model-based average estimator. These approaches are evaluated for their efficacy in regressing missing Doppler velocity log beams when two beams are unavailable. In our study, we used data recorded by a DVL mounted on an autonomous underwater vehicle operated in the Mediterranean Sea. We found that both deep learning architectures outperformed model-based approaches by over 16\% in velocity prediction accuracy.

\end{abstract}


\section{Introduction}
\noindent
An autonomous underwater vehicle (AUV) is a self-propelled, untethered robot designed to operate underwater without direct human control. These vehicles are equipped with sensors, navigation systems, and sometimes manipulators, enabling them to perform various tasks such as oceanographic research, underwater mapping, pipeline inspection, environmental monitoring, and search and rescue operations. AUVs are valuable tools for exploring and studying the ocean's depths, offering efficiency, flexibility, and the ability to access areas that may be difficult or dangerous for manned vehicles \cite{sahoo2019advancements}. \\
\noindent
The primary navigation sensors equipped in AUVs are the inertial navigation system (INS), which provides a high-rate navigation solution characterized by accumulating error over time, and the Doppler velocity log (DVL), a low-rate sensor that offers accurate velocity updates to mitigate error accumulation \cite{groves2015principles,leonard2016autonomous}.
Navigating through the underwater domain poses significant challenges, particularly for AUVs, whose sole purpose is to accomplish this task. Addressing these challenges is paramount for the successful operation of such vehicles. One of the challenges in underwater navigation is the inability of global navigation satellite system (GNSS) signals to penetrate water, rendering them unreliable for underwater navigation. As a result, the DVL becomes the sole sensor relied upon for correcting the inertial navigation errors. In real-world scenarios, such as passing over trenches in the seabed, encountering obstacles that obstruct the sensor's view, or executing extreme maneuvers, the DVL may fail to operate correctly. This is because its mechanism relies on four beams being transmitted and reflected back to the sensor to estimate speed. In situations where fewer than three beams are reflected back, accurate speed estimation becomes impossible \cite{brokloff1994matrix}.\\
\noindent
Since incomplete DVL measurements pose a critical problem for AUV navigation, several solutions have been proposed in the literature to address this challenge. In \cite{tal2017inertial,eliav2018ins}, an extended loosely coupled (ELC) strategy is introduced as a solution for this circumstance. By utilizing the restricted DVL beam measurements along with external data, the ELC approach computes the three-dimensional velocity of the AUV and integrates it into the navigation filter. Liu \emph{et al.} \cite{liu2018ins} proposed a method grounded in the tightly coupled approach for INS/DVL integrated navigation. This method directly incorporates DVL beam measurements, bypassing the need to transform them into 3-D velocity. It incorporates limited DVL beam measurements alongside depth updates obtained from a pressure sensor. In \cite{karmozdi2020implementation} Initially, the vehicle's translational model is deduced, capturing the dynamics of translational velocity across three directions in the body frame. The integration of this model produces the vehicle's translational velocity in the body frame. This velocity, referred to as pseudo-DVL, functions to offset the lack of DVL in INS/DVL navigation systems. Additional solutions for scenarios involving complete DVL outages were introduced in \cite{klein2020continuous,klein2022estimating}.\\
\noindent
As hardware capabilities advanced to handle heavy computational loads, the utilization of machine learning (ML) and deep learning (DL) techniques emerged to enhance AUV navigation \cite{cohen2023inertial}. In \cite{cohen2022beamsnet}, a DL approach was applied to enhance DVL velocity estimation, while in \cite{or2022hybrid,cohen2024kit}, data-driven methods were employed to estimate the process noise covariance in INS/DVL fusion. In situations of partial or complete DVL outage, various ML and DL approaches have been explored to address this challenge and maintain INS/DVL solutions \cite{ansari2019pseudo,lv2020underwater,wang2021virtual,yao2022virtual,zhu2022hybrid}. Specifically, Yona \emph{et al.} proposed a deep learning approach based on long short-term memory (LSTM) networks in \cite{yona2021compensating,yona2023missbeamnet}. This method, known as MissBeamNet, aims to regress the missing beams using past DVL measurements. Furthermore, in \cite{cohen2022beamsnet,cohen2022libeamsnet,cohen2023set}, a series of architectures called BeamsNet were introduced to tackle scenarios involving complete, partial, and absent DVL measurements. Notably, LiBeamsNet utilized a one-dimensional convolutional neural network (1DCNN) to regress two missing beams.\\
\noindent
In this study, we undertake a comparative analysis between LiBeamsNet and MissBeamNet, alongside a model-based average estimator, to ascertain whether DL approaches surpass model-based methods and whether long-term dependencies offer superior solutions in LSTM networks compared to the feature extraction from nearby windows used in a 1DCNN. We evaluate these strategies using real-world recorded data from an AUV experiment conducted in the Mediterranean Sea, which is available online at \url{https://github.com/ansfl/A-KIT}.\\
\noindent
The remainder of the paper is structured as follows: Section \ref{PF} formulates the problem, while Section \ref{CS} introduces and discusses the competitive strategies. Section \ref{Data} provides an overview of the dataset used in the study, and Section \ref{res} presents the results of the comparison. Finally, Section \ref{con} discusses the conclusions drawn from the results.

\section{Problem Formulation}\label{PF}
\begin{figure}[htbp]
	\centering
		\includegraphics[width=\columnwidth]{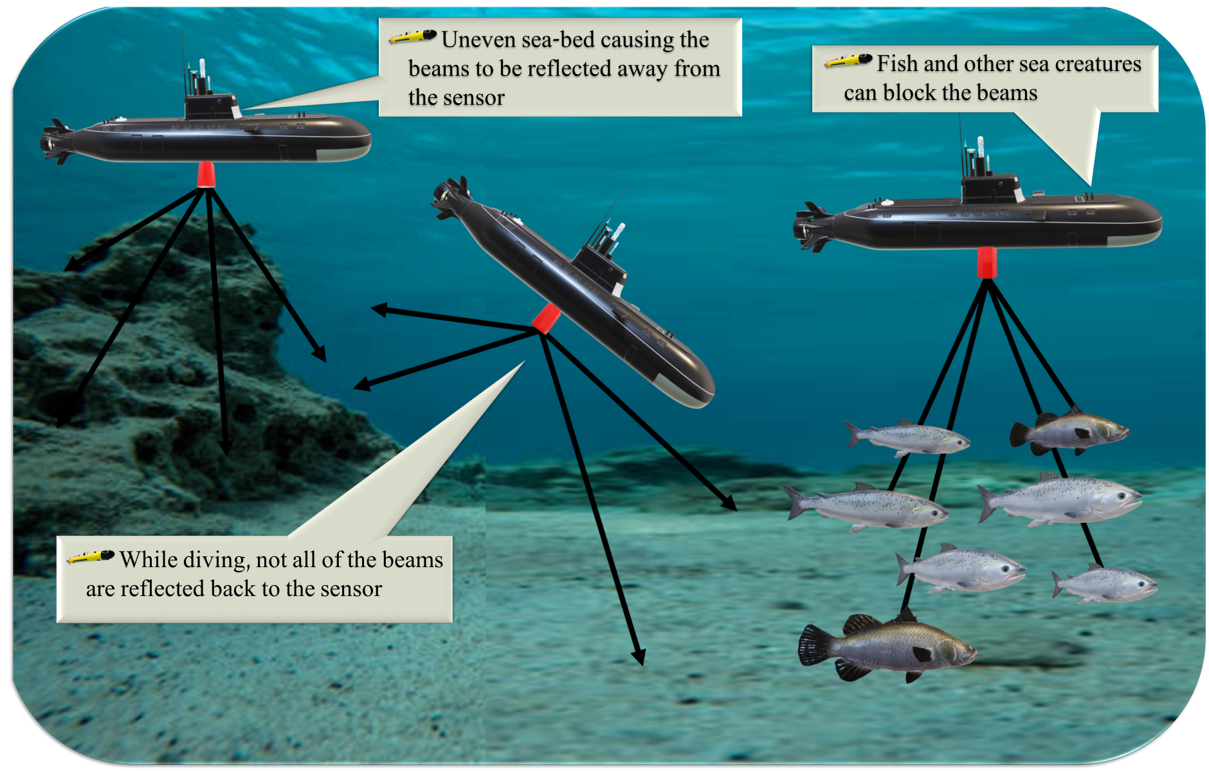}
	  \caption{Illustration of scenarios where two or more DVL beams may be unavailable.}\label{Fig:1}
\end{figure}

\noindent
The DVL operates by transmitting four acoustic beams to the sea floor, and once they are reflected back to the sensor, it calculates the frequency shift, thereby determining the velocity of the AUV in the beam directions. Given that the configuration of the transducers is known and provided by the manufacturer, solving a least squares problem is necessary to obtain the velocity in the DVL frame. The DVL velocity vector estimation based on the beam measurements are \cite{cohen2022beamsnet}:
\begin{equation}\label{eqn:1}
    \centering \boldsymbol{v}_{\text{beam}} = \mathbf{T}_{\text{beam}} \boldsymbol{v}^{d}, \; \mathbf{T}_{\text{beam}} =
        \begin{bmatrix} 
        c\psi_{\text{beam}_1}s\alpha & s\psi_{\text{beam}_1}s\alpha & c\alpha\\
        c\psi_{\text{beam}_2}s\alpha & s\psi_{\text{beam}_2}s\alpha & c\alpha\\
        c\psi_{\text{beam}_3}s\alpha & s\psi_{\text{beam}_3}s\alpha & c\alpha\\
        c\psi_{\text{beam}_4}s\alpha & s\psi_{\text{beam}_4}s\alpha & c\alpha
        \end{bmatrix}
\end{equation} 
\begin{equation}\label{eqn:2}
    \centering
        \psi_{\text{beam}_{\dot{\imath}}} = (\dot{\imath}-1)\cdot 90^{\circ} + 45^{\circ}, \; \dot{\imath}=1,2,3,4.
\end{equation}
\begin{equation}\label{eqn:3}
    \centering
        \boldsymbol{y} = \mathbf{T}_{\text{beam}} [\boldsymbol{v}^{d} (\boldsymbol{1} + \boldsymbol{s}_{\text{DVL}})] + \boldsymbol{b}_{\text{DVL}} + \boldsymbol{n}
\end{equation}
\begin{equation}\label{eqn:4}
    \centering
        \hat{\boldsymbol{v}}^{d} =
        \underset{\boldsymbol{v}^{d}}{\argmin}{\| \boldsymbol{y} - \mathbf{T}_{\text{beam}} \boldsymbol{v}^{d} \|}^{2}.
\end{equation} 

\begin{equation}\label{eqn:5}
    \centering
        \hat{\boldsymbol{v}}^{d} = (\mathbf{T}_{\text{beam}}^{T} \mathbf{T}_{\text{beam}})^{-1} \mathbf{T}_{\text{beam}}^{T} \boldsymbol{y}.
\end{equation}
\noindent
In \eqref{eqn:1}, $s$ and $c$ represent $\sin$ and $\cos$, respectively, $\boldsymbol{v}_{\text{beam}} \in \mathbb{R}^{4\times1}$ denotes the velocity in the beam directions, and $\boldsymbol{v}^{d} \in \mathbb{R}^{3\times1}$ is the velocity in the DVL frame. Each beam transducer undergoes a rotational transformation with a yaw angle $\psi_{\text{beam}}$ and a pitch angle $\alpha$, where $\alpha$ is a constant predefined by the manufacturer and $i$ is the beam index. The measurements, prone to inherent errors, are modeled in \eqref{eqn:3}, with $\boldsymbol{b}_{\text{DVL}} \in \mathbb{R}^{4\times1}$ as the bias vector, $\boldsymbol{s}_{\text{DVL}} \in \mathbb{R}^{3\times1}$ as the scale factor vector, and $\boldsymbol{n} \in \mathbb{R}^{4\times1}$ representing zero-mean white Gaussian noise. Upon obtaining raw measurements, the subsequent step involves extracting $\boldsymbol{v}^{d}$ by filtering the data according to the cost function in \eqref{eqn:4}. The solution of \ref{eqn:4} is the velocity in the DVL frame as detailed in \eqref{eqn:5}.\\
\noindent
In real-world scenarios, the DVL may face challenges leading to the reflection of acoustic beams in directions other than toward the sensor. When two or more beams are unavailable, the DVL cannot perform the task of estimating velocity. Examples of such scenarios may include passing over trenches on the sea floor, sea creatures obstructing the view of the sensor, and extreme roll and pitch maneuvers, including the AUV's initial operation, diving. Illustrations of these scenarios are visualized in Fig. \ref{Fig:1}.
In these cases, the navigation solution relies solely on the INS, which is known for accumulating errors over time and cannot serve as a standalone sensor for navigation.
\section{Competitive Strategies}\label{CS}
\noindent
To address incomplete DVL measurements, we focused on scenarios where two beams are missing, and as a consequence, the velocity vector cannot be estimated. We examine two state-of-the-art deep learning architectures that employ different techniques yet utilize only DVL information. The first one is LiBeamsNet, which employs 1D CNN on past complete DVL measurements combined with the current observation of the partial beam measurements to regress the missing beams. The second one is MissBeamNet, which uses LSTM architecture to regress the missing beams given past measurements and the current observation.
Both approaches demonstrated good performance in estimating the missing beams. However, a question arises regarding which of them performs better: the 1DCNN architectures that work by extracting features from the window of past measurements or the LSTM network that considers larger time dependencies.
\begin{figure}[htbp]
	\centering		
        \includegraphics[width=\columnwidth]{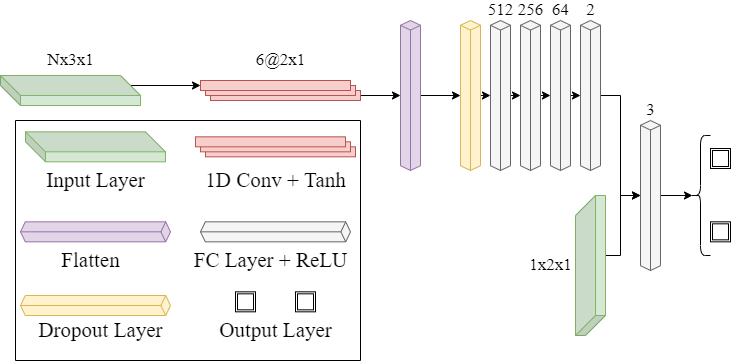}
	  \caption{The LiBeamsNet architecture, as introduced in \cite{cohen2022libeamsnet}, presenting a 1DCNN for extracting features from $N$ past DVL beam measurements.}\label{LiBeams}
\end{figure}
\begin{figure}[htbp]
	\centering		
        \includegraphics[width=\columnwidth]{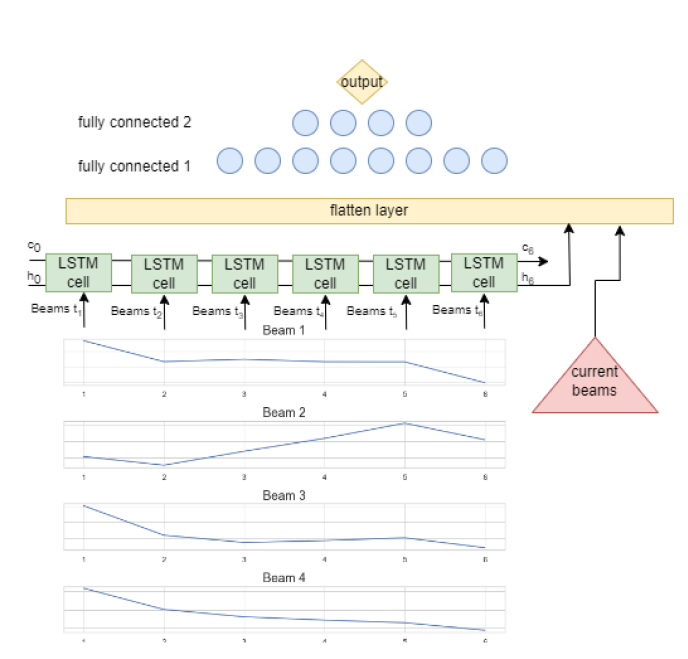}
	  \caption{The MissBeamNet architecture, as described in \cite{yona2023missbeamnet}, employs an LSTM structure to analyze $N$ past DVL beam measurements, focusing on capturing longer-term dependencies.}\label{MissBeam}
\end{figure}

\noindent
Both networks underwent training for 100 epochs with a learning rate of 0.001, a learning rate decay of factor 0.1 after 50 epochs, a mini-batch size of 4, and the ADAM optimizer. In addition, for both networks, $N=3$ past DVL measurements were employed.\\
\noindent
In LiBeamsNet, the measurements undergo processing through a 1D convolutional layer comprising six filters of size $2\times1$ aimed at extracting features from the data. Following this, the features extracted are flattened and then subjected to a dropout layer with a dropout rate of 0.2. Subsequently, a sequence of fully connected layers is employed, and the current partial beam measurement is integrated before passing through the last fully connected layer, ultimately producing a $4\times1$ vector representing the estimated DVL four-beam velocity. The architecture and activation functions after each layer are illustrated in Fig. \ref{LiBeams}. In MissBeamNet, after the LSTM with a hidden size of 500 performs feature extraction, these features are concatenated with the available beam measurements into a fully connected layer. This layer executes the final processing, resulting in the output of the regressed missing beams, as can be seen in Fig. \ref{MissBeam}. For comparison, we examined a conventional average estimator, which takes the average of the $N=3$ past DVL beam measurements to regress the missing ones.
\section{Dataset}\label{Data}
Experiments were conducted in the Mediterranean Sea near the shores of Haifa, Israel, utilizing the Snapir AUV to collect data. Snapir is a modified ECA Group A18D mid-size AUV designed for deep-water applications, capable of autonomous missions up to 3000 meters in depth and featuring a 21-hour endurance \cite{ECA_AUV}. It is equipped with the Teledyne RDI Work Horse Navigator DVL \cite{TeledyneMarine_DVL}, which achieves precise velocity measurements with a standard deviation of $0.02\;[\text{m/s}]$ and operates at a frequency of $1\;[\text{Hz}]$.\\
\noindent
The dataset was recorded in June 2022, and the train set consists of eleven distinct data sections, each lasting $400 \;[\text{sec}]$, featuring diverse dynamics for comprehensive training. To assess the proposed approach, two additional segments of data, each lasting $400\;[\text{sec}]$ and not included in the training set, were examined as the test set. To emulate a unit under test, we implemented the error model \eqref{eqn:3} with specific parameter values: $\alpha = 20^{\circ}$, $\textbf{s}_{DVL} = \textbf{0}_{3\times1}$, $\textbf{b}_{DVL} =0.001\cdot \textbf{1}_{3\times1}; [m/s]$, and $\textbf{n}$ follows a normal distribution with zero mean and a standard deviation of $0.001\;[m/s]$. All the data is accessible on GitHub at the following link: \url{https://github.com/ansfl/A-KIT}, as introduced in \cite{cohen2024kit}.
\section{Results}\label{res}
\begin{figure}[htbp]
	\centering		
        \includegraphics[width=\columnwidth]{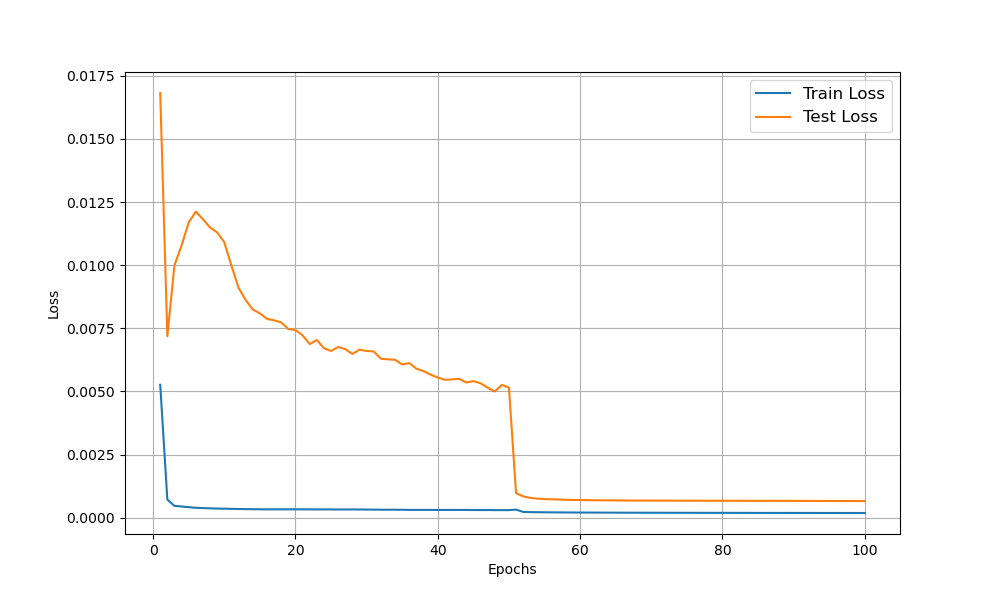}
	  \caption{The loss values of both the train and test sets for LiBeamsNet as a function of the epoch number.}\label{fig:lossLibeamsnet}
\end{figure}

\begin{figure}[htbp]
	\centering		
        \includegraphics[width=\columnwidth]{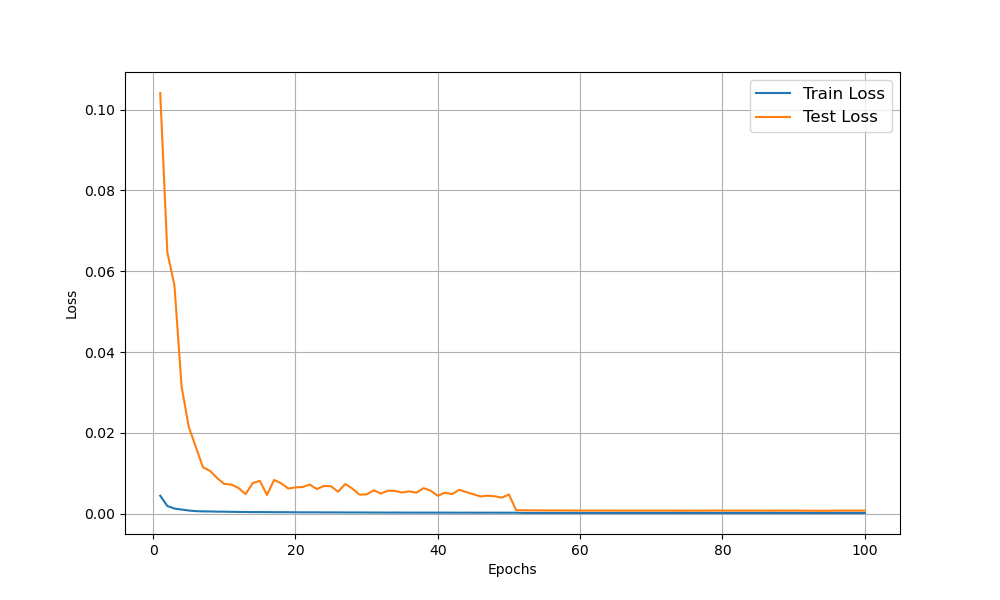}
	  \caption{The loss values of both the train and test sets for MissBeamNet as a function of the epoch number.}\label{fig:lossmissbeamsnet}
\end{figure}

\noindent
In the initial phase of the analysis, we examined the loss values as a function of the number of epochs. In Fig. \ref{fig:lossLibeamsnet}, both the training and testing loss values for LiBeamsNet are depicted, while Fig. \ref{fig:lossmissbeamsnet} illustrates the loss values for MissBeamNet. Both graphs show a convergence of the training loss, with LiBeamsNet achieving a lower value. Examining the test loss values reveals that MissBeamNet converges more rapidly and exhibits a monotonically decreasing trend, whereas LiBeamsNet takes a longer time to converge. This suggests that MissBeamNet is better able to generalize the problem.
For a robust evaluation, we considered four matrices: 1) root mean squared error (RMSE), 2) mean absolute error (MAE), 3) the coefficient of determination ($R^{2}$), and 4) the variance accounted for (VAF). RMSE and MAE measure velocity errors in $[m/s]$, while $R^{2}$ and VAF are unitless. The definitions of these matrices are as follows:
\begin{equation}\label{eqn:6}
    \centering
        RMSE(\boldsymbol{x}_{\dot\imath},\hat{\boldsymbol{x}}_{\dot\imath})=\sqrt{\frac{\sum_{\dot\imath=1}^{N}(\boldsymbol{x}_{\dot\imath}-\hat{\boldsymbol{x}}_{\dot\imath})^{2}}{M}}
\end{equation}
\begin{equation}\label{eqn:7}
    \centering
        MAE(\boldsymbol{x}_{\dot\imath},\hat{\boldsymbol{x}}_{\dot\imath})=\frac{\sum_{\dot\imath=1}^{M}|\boldsymbol{x}_{\dot\imath}-\hat{\boldsymbol{x}}_{\dot\imath}|}{M}
\end{equation}
\begin{equation}\label{eqn:8}
    \centering
        R^{2}(\boldsymbol{x}_{\dot\imath},\hat{\boldsymbol{x}}_{\dot\imath})=1- \frac{\sum_{\dot\imath=1}^{M}(\boldsymbol{x}_{\dot\imath}-\hat{\boldsymbol{x}}_{\dot\imath})^{2}}{\sum_{\dot\imath=1}^{M}(\boldsymbol{x}_{\dot\imath}-\bar{\boldsymbol{x}}_{\dot\imath})^{2}}
\end{equation}
\begin{equation}\label{eqn:9}
    \centering
        VAF(\boldsymbol{x}_{\dot\imath},\hat{\boldsymbol{x}}_{\dot\imath})=[1-\frac{var(\boldsymbol{x}_{\dot\imath}-\hat{\boldsymbol{x}}_{\dot\imath})}{var(\boldsymbol{x}_{\dot\imath})}]\times100
\end{equation}
where $M$ denotes the number of samples, $\boldsymbol{x}_{\dot\imath}$ signifies the ground truth velocity vector norm of the DVL, $\hat{\boldsymbol{x}}_{\dot\imath}$ represents the predicted velocity vector norm of the AUV, generated after regressing the two missing beams by the network, $\bar{\boldsymbol{x}}_{\dot\imath}$ denotes the mean of the ground truth velocity vector norm of the DVL, and $var$ stands for variance. It is noteworthy that achieving a VAF of $100$, an $R^{2}$ of $1$, and both RMSE and MAE being zero would characterize the model as exceptional \cite{cohen2022beamsnet}.
\\
\noindent
By looking at Table \ref{tab:1}, we can compare the outputs of the discussed strategies.
\begin{table}[htbp]
\centering
\caption{A table summarizing the findings regarding the desired metrics and illustrating the improvement achieved by the DL approaches compared to the model-based average estimator.}
\label{tab:1}
\resizebox{\columnwidth}{!}{%
\begin{tabular}{|c|c|c|c|}
\hline
Metric         & LiBeamsNet & MissBeamNet & Average \\ \hline
RMSE {[}m/s{]} & 0.0653     & 0.0662      & 0.0794  \\ \hline
RMSE {[}\%{]}  & 17.75      & 16.62       & N/A     \\ \hline
MAE {[}m/s{]}  & 0.0451     & 0.0422      & 0.0558  \\ \hline
MAE {[}\%{]}   & 19.12      & 24.37       & N/A     \\ \hline
$R^{2}$        & 0.9899     & 0.9896      & 0.9850  \\ \hline
VAF            & 99.11      & 99.04       & 98.55   \\ \hline
\end{tabular}%
}
\end{table}
It is evident that the LiBeamsNet architecture, utilizing the 1DCNN strategy, outperforms MissBeamNet in every metric, albeit by a marginal margin that could be attributed to statistical fluctuations, resulting in approximately similar outputs. In terms of RMSE and MAE, a significant improvement of over 16\% was observed with the adoption of the data-driven approaches compared to the model-based average estimator. Analyzing the statistical properties $R^{2}$ and VAF, all three approaches exhibit a good fit for the given task. Nevertheless, the data-driven methods consistently demonstrate superior performance over the model-based average estimator.
\section{Conclusions}\label{con}
AUVs are specifically engineered to function in the depths of the underwater environment beyond the reach of human exploration. Typically, AUV navigation heavily relies on data from DVL sensors to rectify any accumulated errors from the INS. However, in situations where the DVL fails to provide accurate velocity updates, which could occur due to various scenarios, the AUV may be compelled to abort its mission, subject to the manufacturer's protocol.
\\ \noindent
One common approach to tackle this challenge is to forecast the missing beam measurements using an average estimator, which utilizes past measurements. In our study, we compare two state-of-the-art data-driven methods, LiBeamsNet and MissBeamNet, with a model-based average estimator. Our results reveal that despite employing distinct DL architectures—one based on 1DCC and the other on LSTM—both exhibit similar performance, with a statistically insignificant difference, yet outperforming the average estimator by more than 16\% in terms of velocity prediction accuracy.
\section*{Acknowledgments}
\noindent
N.C. is supported by the Maurice Hatter Foundation and
University of Haifa presidential scholarship for outstanding students on a direct Ph.D. track.
\bibliographystyle{ieeetr}
\bibliography{refs}

\end{document}